\documentclass[11pt]{article}
\usepackage[preprint]{acl}
\usepackage{times}
\usepackage{latexsym}
\usepackage[T1]{fontenc}
\usepackage[utf8]{inputenc}
\usepackage{microtype}
\usepackage{inconsolata}
\usepackage{graphicx}
\usepackage{booktabs}
\usepackage{amsmath}
\usepackage{amssymb}
\usepackage{multirow}
\usepackage{colortbl}
\definecolor{oursrow}{HTML}{EAF2FB}
\usepackage{tcolorbox}
\tcbuselibrary{skins,breakable}
\definecolor{pframe}{HTML}{4A6FA5}\definecolor{pback}{HTML}{F7F9FC}\definecolor{ptitle}{HTML}{DCE6F2}
\definecolor{eframe}{HTML}{5B8C5A}\definecolor{eback}{HTML}{F6FAF5}\definecolor{etitle}{HTML}{E1EEDF}
\tcbset{apxbox/.style={enhanced,breakable,boxrule=0.5pt,arc=2pt,left=4pt,right=4pt,top=2pt,bottom=3pt,
  toptitle=1.5pt,bottomtitle=1.5pt,coltitle=black,fonttitle=\small\bfseries,fontupper=\small,
  before skip=6pt,after skip=6pt}}
\newtcolorbox{promptbox}[1]{apxbox,colframe=pframe,colback=pback,colbacktitle=ptitle,title={#1}}
\newtcolorbox{expbox}[1]{apxbox,colframe=eframe,colback=eback,colbacktitle=etitle,title={#1}}
\newcommand{\pfield}[1]{\par\noindent\textbf{#1:}\ }
\usepackage{enumitem}

\title{ImageRSI: Verification-Driven Recursive Self-Improvement for Agentic Image Generation}
\author{
  Shuo Han$^{1,2,*}$\quad
  Zhilin Wang$^{2,*}$\quad
  Peng Ling$^{2}$\quad
  Wenjia Gong$^{2}$\\
  \textbf{Yang Liu}$^{2}$\quad
  \textbf{Guanjun Jiang}$^{2}$\quad
  \textbf{Yukun Cao}$^{3,\dagger}$\quad
  \textbf{Xike Xie}$^{1,\dagger}$\\[4pt]
  $^{1}$University of Science and Technology of China\\
  $^{2}$Qwen Business Unit of Alibaba, Alibaba Token Hub\\
  $^{3}$Xidian University\\
  \small\texttt{shuo.han@mail.ustc.edu.cn},
  \small\texttt{\{wzl446229,lingpeng.lp,wenjia.gwj,zhuda.zd,guanj.jianggj\}@alibaba-inc.com},\\
  \small\texttt{caoyukun@xidian.edu.cn},
  \small\texttt{xkxie@ustc.edu.cn}\\[2pt]
  \small\textnormal{$^{*}$Equal contribution.\quad $^{\dagger}$Corresponding authors.}
}

\begin{document}
% Company logo at the top of the first page (arXiv version only).
\AddToHookNext{shipout/foreground}{%
  \put(2.5cm,-1.1cm){%
    \makebox(0,0)[lt]{%
      \includegraphics[height=0.95cm,keepaspectratio]{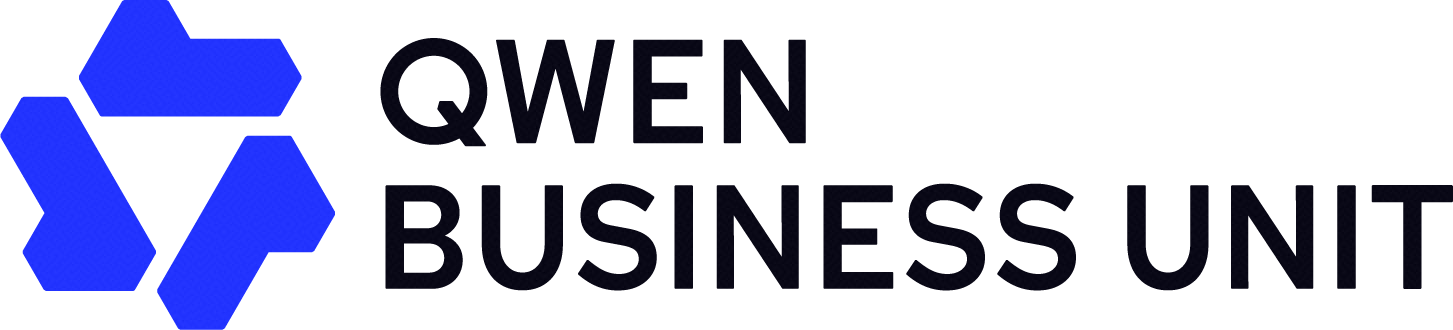}%
        
    }%
  }%
}
\maketitle

% ============================================================
%  ABSTRACT
% ============================================================
\begin{abstract}
Complex text-to-image requests remain difficult to satisfy reliably. Verifying and repairing an image can fix an individual output, but a successful repair does not necessarily carry over to other requests. We present \textbf{ImageRSI}, a framework that keeps the image generator frozen and combines three components: five-dimensional verification and repair within each task, an experience library shared across tasks, and offline updating of the judge. Candidate experiences enter the library only after paired A/B validation, and the judge is updated with supervised fine-tuning and GRPO on accumulated image--judgment records and evaluated before reuse. We also introduce \textbf{ImageReq-Bench}, a benchmark of 961 requests built around hard cases in real-world image-generation requests, each paired with an evaluation checklist. ImageRSI achieves a Checklist Strict Accuracy (CSA) of 52.6\% on MindBench (with Nano Banana Pro as the generator) and 65.66\% on ImageReq-Bench, and we compare system outcomes across judge-training configurations.
\end{abstract}

% ============================================================
%  1  INTRODUCTION
% ============================================================

\section{Introduction}
\label{sec:intro}

Text-to-image models such as GPT-Image and Wan2.7-Image~\citep{mao2026wan} generate high-quality images with fine-grained style control and strong instruction following. Yet requests that combine multiple constraints still reveal failures in factual knowledge, exact text rendering, object counts, attribute binding, and spatial relations \cite{huang2023t2i,zhang2025strict,sun2026t2i}. Requests may also leave some requirements implicit. For example, a request for a labeled diagram of a solar eclipse implicitly requires the Moon to appear between the Sun and Earth, with the three bodies aligned and legible labels identifying each body. Reliable generation therefore requires inferring the intended constraints, verifying that the resulting image satisfies them, and applying targeted repair when needed.

Beyond fixing the current image, verification and repair leave trajectories that record the errors, the changes made to the generation conditions, and the resulting outcomes \cite{ye2025genpilot, yang2024idea2img,zhu2025intelligent}; these trajectories can provide candidate strategies for future tasks. However, a single successful repair does not establish that a strategy is reliable or transferable: stochastic generation can produce improvements by chance, redundant trajectories can overstate the evidence, and judge errors can reinforce incorrect diagnoses. Candidate experience therefore requires validation, explicit conditions of applicability, and criteria for revising or retiring it.

Existing agentic systems address these issues only in part. Some apply verification and iterative repair to individual requests~\cite{shen2026agentic,chen2025t2i,ouyang2026consistency,he2026omnivchat,feng2026gen}, and additionally incorporate cross-task memory~\citep{qwen2026imageagent,xiong-etal-2026-memory}. Retaining feedback, however, does not determine when the resulting experience is actually useful. Evaluation faces a related limitation: fixed-prompt benchmarks~\citep{niu2025wise,yang2026mindbench} measures final-output quality, but it cannot by itself establish whether an agent improves as it accumulates experience across tasks.

\begin{figure}[t]
    \centering
    \includegraphics[width=\columnwidth]{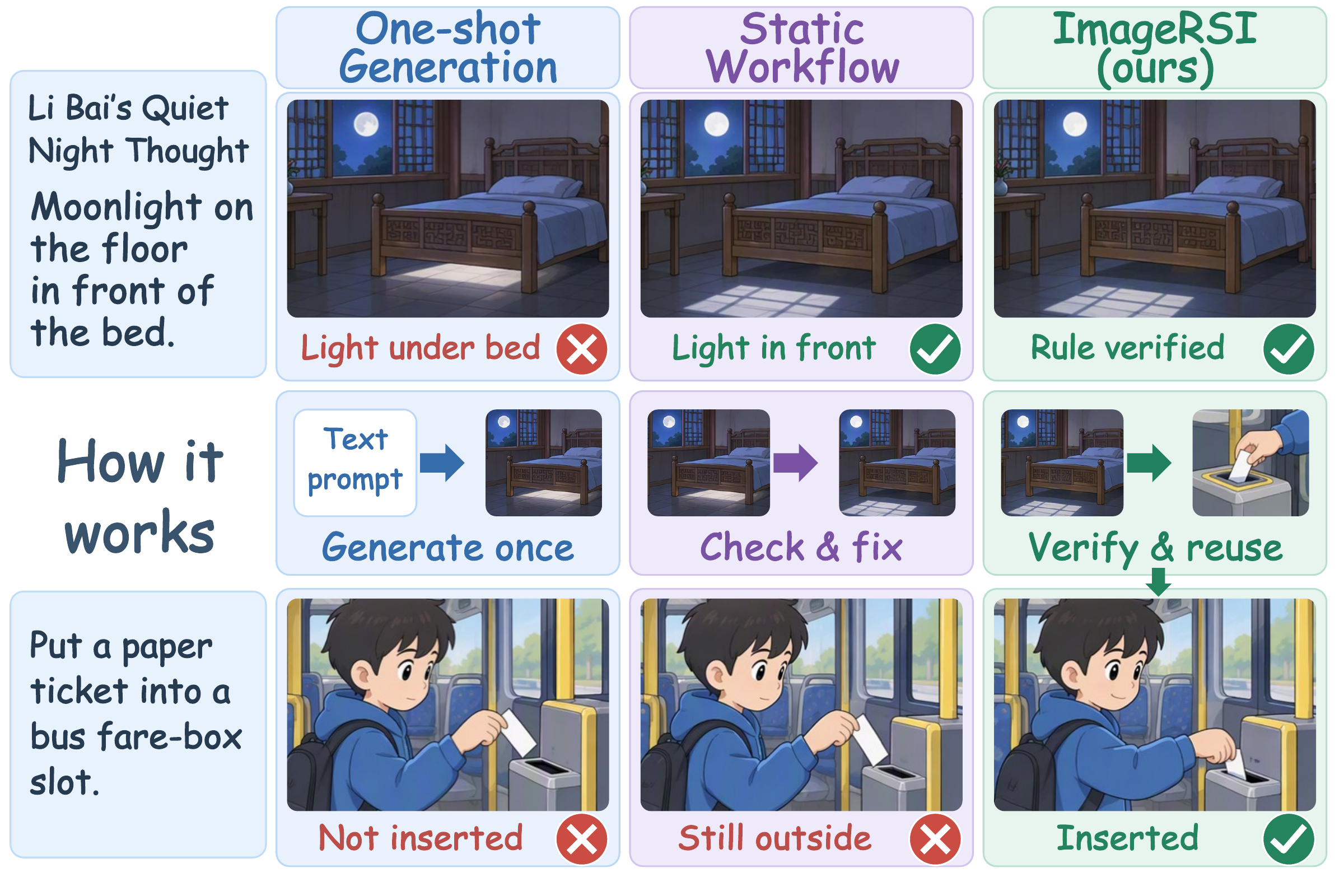}
    \caption{Schematic illustration of task-level repair and cross-task reuse of a relation-grounding strategy.}
    \label{fig:figure1}
\end{figure}

We propose \textbf{ImageRSI}, a verification-driven recursive self-improvement framework that couples within-task repair, cross-task experience reuse, and offline judger self-evolution, where the judger is the verification model that scores images. Within each task, the agent parses the request, retrieves relevant experience and, when needed, external knowledge, generates an image, and evaluates it for Alignment, Faithfulness, Quality, Aesthetic, and Physics. Defect feedback then guides targeted prompt repair while leaving the original requirements unchanged. Across tasks, an experience library stores candidate strategies, each recording when it applies and when it does not, the prompt change it makes, and its observed effects and side effects; paired A/B validation decides which experiences are admitted for reuse. For judger self-evolution, selected trajectory images are paired with model-generated score-and-reason annotations for supervised fine-tuning (SFT), followed by GRPO  \cite{shao2024deepseekmath} to optimize scoring behavior. After evaluation, a candidate judge can guide verification and repair in a later collection round, yielding new records for further updates. Experience admission updates inference-time knowledge, whereas judger self-evolution updates the judge's parameters offline; the image generator stays frozen throughout. Figure~\ref{fig:figure1} gives a schematic example of task-level repair and cross-task reuse, and Figure~\ref{fig:framework} illustrates the three processes in detail.

\noindent Our contributions are as follows:
\begin{itemize}
\item We propose ImageRSI, a verification-driven recursive self-improvement framework for agentic image generation with a frozen image generator. Within each task, five-dimensional verification localizes unmet requirements, and repair rewrites the prompt to target these defects while locking the parsed requirements, so that satisfied requirements are preserved across rounds.
\item We formulate two complementary uses of repair trajectories: an experience library records when each strategy applies and requires paired A/B validation before reuse, while accumulated image--judgment records support offline judge post-training and evaluation before redeployment.
\item We introduce ImageReq-Bench, a benchmark of 961 complex image-generation requests built around hard cases in real-world image-generation requests. Each request is linked to its source records and an evaluation checklist for assessing whether its requirements are jointly satisfied.
\item ImageRSI achieves state-of-the-art CSA on both benchmarks, reaching 52.6\% on MindBench and 65.66\% on ImageReq-Bench and surpassing the strongest baselines by 10.2 and 9.28\,pp, respectively.
\end{itemize}

% ============================================================
%  2  RELATED WORK
% ============================================================

\section{Related Work}
\label{sec:related}

\paragraph{Agentic image generation.}
Diffusion and multimodal generators have substantially advanced image synthesis~\citep{rombach2022ldm,podell2024sdxl,esser2024sd3,sun2024emu,team2024chameleon}. However, satisfying factual, textual, spatial, and physical requirements that interact with one another remains challenging, as examined in MindBench and WISE~\citep{yang2026mindbench,niu2025wise}. Agentic approaches handle complex requests through planning, external tools, and iterative evaluation~\citep{xue2024genagent,yang2026mindbench,qwen2026imageagent}, and Qwen-Image-Agent already reuse memory across tasks~\citep{qwen2026imageagent}. ImageRSI additionally specifies an explicit validation protocol and applicability conditions for candidate experience.

\paragraph{Experience-driven improvement.}
Prior work has used verification and search to improve reasoning~\citep{hosseini2024vstar,zhang2024restmcts,cao2025graphinsight,NEURIPS2025_5c882988}, distilled lessons from agent trajectories~\citep{zhao2024expel}, and explored iterative improvement of programs or agent context~\citep{zelikman2024stop,zhang2026dgm,zhang2026agentic,yang2026autoskill,wei2026dynamic,wu-etal-2026-rsmem,zheng2026dream}. ImageRSI brings these ideas to image generation in two ways: it admits reusable experience through A/B testing, and it supports judger self-evolution through offline post-training on versioned verification records. Experience validation checks whether a piece of experience applies before it is activated, and the judge-training loop turns newly collected failures and successes into supervision for later versions.

% ============================================================
%  3  METHOD
% ============================================================
\section{Method}
\label{sec:method}
 
\subsection{Overview}
\label{sec:framework}

ImageRSI uses image verification to repair the current output, maintain experience for later tasks, and accumulate training records for offline judger self-evolution. These processes address three connected challenges: identifying unmet requirements, deciding when a repair strategy can be reused, and adapting the judge to observed failures.

\begin{figure*}[t]
\centering
\includegraphics[width=\textwidth]{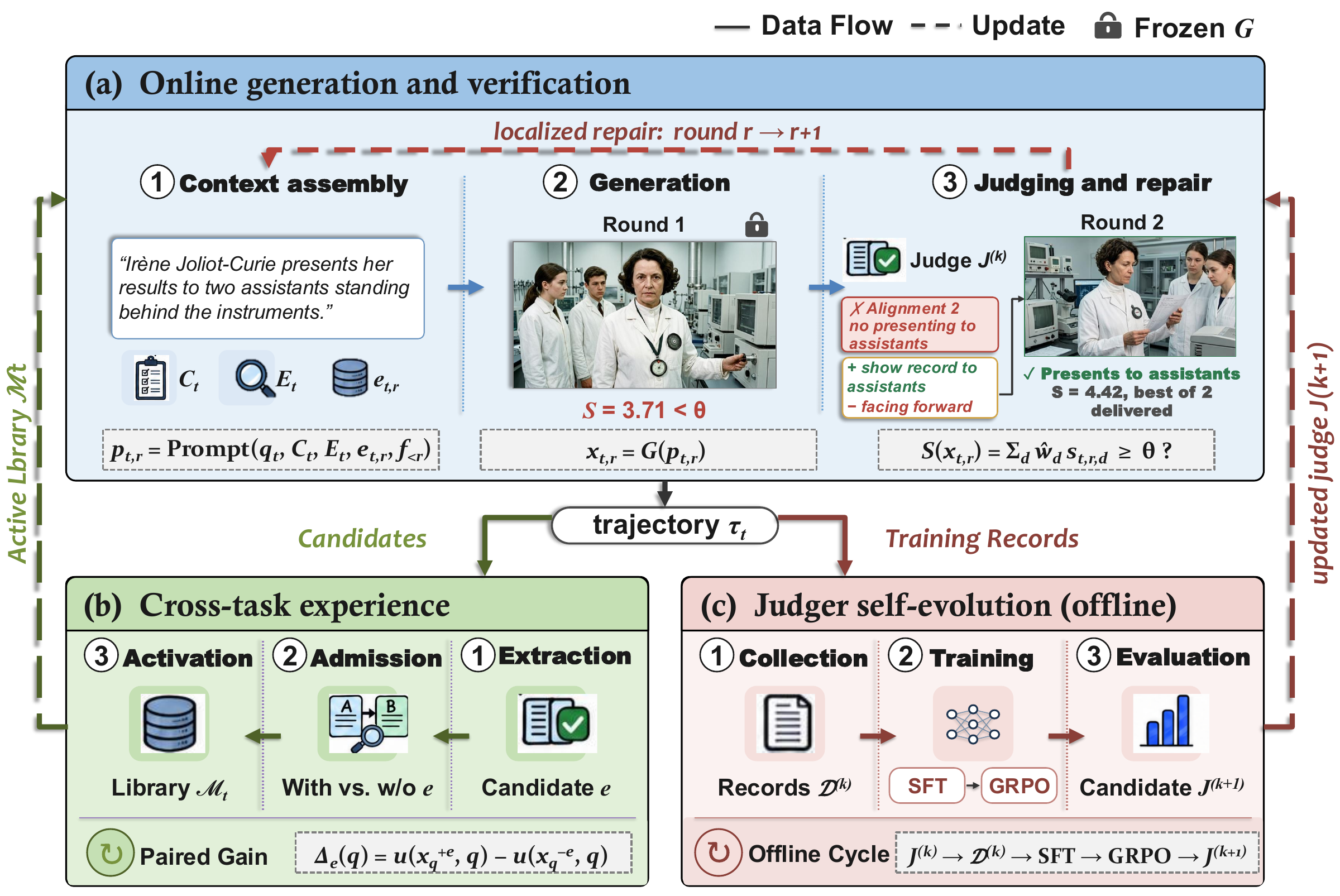}
\caption{Overview of ImageRSI. (a)~Online generation and verification: the frozen generator $G$ produces an image, and the judge $J^{(k)}$ verifies it and guides targeted repair. (b)~Cross-task experience: candidates extracted from trajectories are admitted by paired A/B validation and activated for later tasks. (c)~Offline judger self-evolution: trajectory records train a candidate judge with SFT and GRPO, which replaces $J^{(k)}$ after evaluation. Solid and dashed arrows denote data flow and updates, respectively.}
\label{fig:framework}
\end{figure*}

For task $t$, let $q_t$ denote the request, $i_t$ an optional input image, and $\mathcal{M}_{t-1}$ the available experience library. Given a frozen image generator $G$ and the judge $J^{(k)}$ from training iteration $k$, the task-level workflow produces an image $x_t$ and a trajectory $\tau_t$ that records the generation conditions, verification results, and repair actions. As shown in Figure~\ref{fig:framework}, this trajectory provides both candidate experience for $\mathcal{M}$ and supervision for a later judge version. Experience updates the inference-time context, whereas judger self-evolution updates the judge's parameters between data-collection rounds. The generator stays frozen throughout each run.

\subsection{Online Generation and Verification}
\label{sec:query_rewrite}

\paragraph{Requirement-preserving generation.}
\label{sec:req_parse}
The workflow first parses $q_t$ into a set of requirements $C_t=\{c_{t,1},\ldots,c_{t,m}\}$, covering entities, attributes, counts, spatial relations, exact text, layout, and image-to-image (I2I) preservation or exclusion conditions. Holding $C_t$ fixed across repair rounds gives a stable target: a revision must satisfy an unmet requirement rather than drop it from the evaluation criteria.

\label{sec:exp_recall}
\label{sec:retrieval}
Generation context combines factual and visual evidence, gathered through bounded ReAct-style retrieval~\citep{yao2022react}, with procedural advice drawn from active experience. A language model selects the relevant entries from a compact metadata index and adapts their actions to the current request (Appendix~\ref{app:experience_records}).

At round $r$, the prompt combines the request, the fixed requirements, the available evidence, the selected experience, and feedback from earlier rounds. After a failed round, experience selection is rerun with the observed defects as additional context, and the image generator produces one image from the resulting conditions. Appendix~\ref{app:prompt_contracts} gives the prompt interfaces for each stage.

\paragraph{Verification and targeted repair.}
\label{sec:judger}
\label{sec:verification}
The judge evaluates five core dimensions: \textbf{Alignment} (satisfaction of explicit instructions), \textbf{Faithfulness} (correct referents and factual content), \textbf{Quality} (visible defects), \textbf{Aesthetic} (visual presentation), and \textbf{Physics} (physical or spatial coherence). For each dimension, the verification interface returns a score, defect evidence, and a confidence value. For trained judges, which do not output confidence, an inference adapter supplies constant values (Appendix~\ref{app:supervision_data}). Deterministic rules convert these judgments into an acceptance decision.

Let $\mathcal{A}_t$ be the set of active dimensions. For each active dimension $d$, let $s_{t,r,d}$ denote the score of image $x_{t,r}$ and $a_{t,r,d}$ a confidence adjustment. The aggregate score is then
\begin{equation}
\label{eq:score}
\begin{aligned}
\widehat{w}_{t,r,d}
&=\frac{w_d a_{t,r,d}}{\sum_{j\in\mathcal{A}_t}w_j a_{t,r,j}},\\
S(x_{t,r})
&=\sum_{d\in\mathcal{A}_t}\widehat{w}_{t,r,d}s_{t,r,d}.
\end{aligned}
\end{equation}
An image is accepted only if it meets the score threshold and triggers no hard veto (Appendix~\ref{app:scoring_acceptance}). Scoring is based on the request, the parsed requirements, and the image evidence; a separate attribution step may then examine the generation prompt and the selected experience to propose repairs, without altering the scores.

A separately enabled constraint-aware variant checks each original requirement individually and keeps already-satisfied clauses intact during repair (Appendix~\ref{app:constraint_verification}).

\paragraph{Stopping and delivery.}
The workflow stops at the first accepted round or once its generation budget is exhausted (Appendix~\ref{app:generation_delivery}). If no judged round is accepted, it returns the highest-scoring available image as a fallback.\footnote{External CSA evaluation is separate from in-loop acceptance (Appendix~\ref{app:csa_protocol}).} Trajectories are retained for both accepted and fallback outputs.

\subsection{Cross-Task Experience}
\label{sec:experience}

A correction observed on one task may depend on the specifics of that task or on variability in generation. The experience protocol therefore separates \emph{candidate extraction} from \emph{admission for reuse}.

\paragraph{From trajectories to candidate strategies.}
The extractor proposes a procedural strategy by examining observed defects, prompt changes, and post-repair outcomes. Each candidate strategy records a \textbf{Trigger}, \textbf{Negative Trigger}, \textbf{Action}, \textbf{Effect}, and \textbf{Known Side Effects}. A versioned experience library links each entry to its retrieval metadata, source trajectories, and trial records. Appendices~\ref{app:experience_records} and~\ref{app:operator_examples} detail the record fields and representative entries.

\paragraph{Paired admission and maintenance.}
For a candidate $e$, the validation protocol compares images generated with and without its action, keeping the request, upstream inputs, generator configuration, and seed fixed. A blinded evaluator scores the resulting image pair. For validation task $q$, the recorded difference is
\begin{equation}
\label{eq:experience_delta}
\Delta_e(q)=u(x_q^{+e},q)-u(x_q^{-e},q),
\end{equation}
where $u$ is the score assigned by the paired evaluator, which is distinct from the in-loop aggregate $S$. Calibration tests whether eligible trials support reuse, and a separate holdout phase tests the candidate before it is activated. Both phases account for improvements as well as regressions, following the admission criteria in Appendix~\ref{app:admission}.

The admission protocol promotes a candidate version to the active library after calibration and holdout. Trigger and Negative Trigger conditions constrain when these versions are later selected, and recorded outcomes inform their revision, consolidation, or retirement. Unless experience evolution is explicitly enabled, evaluation uses a fixed snapshot of the library.

\subsection{Judger Self-Evolution}
\label{sec:judge_evolution}

Judger self-evolution updates the judger, i.e., the verification model that serves as the in-loop judge, using records accumulated during generation and repair. The judge remains fixed within a task; parameter updates take place offline between collection rounds. This process converts task-level observations into supervision for later judge versions, whose five-dimensional scores and evidence-based explanations support acceptance decisions and prompt repair.

\paragraph{From trajectories to training records.}
Generation and repair under judge $J^{(k)}$ produce trajectories $\mathcal{T}^{(k)}$. Selected images are paired with score-and-reason annotations to form a versioned training set $\mathcal{D}^{(k)}$. Each input $z=(q,C,E,i,x)$ contains the request, parsed requirements, permitted evidence, optional task input image, and generated image; its target $y$ contains a score and an evidence-based reason for each dimension. Targets come from model-generated annotations; acceptance by $J^{(k)}$ is not treated as ground truth. For example, an image that omits requested text can receive a low Alignment score with an explanation of the omission. Each data version combines retained records with additional or oversampled cases addressing observed scoring weaknesses (Appendix~\ref{app:supervision_data}).

\paragraph{Post-training the judge.}
SFT trains the judge to predict $y$ from $z$, with loss computed only on response tokens (Appendix~\ref{app:judge_training_objective}). The GRPO stage, trained on requests disjoint from the SFT data, further optimizes scoring behavior while preserving the score-and-reason output format.\footnote{Advanced post-training methods~\cite{liu2026dapo,deng2026latent,han2026see,deng2025llm,feng2026visionweave,ma2026fipo} are orthogonal to our framework and can be naturally integrated. We adopt standard GRPO to keep the optimization protocol general and reproducible.} Its reward combines pointwise assessment of the five scores and their rationales with a within-group ranking adjustment; severe false accepts and critical fabrications or omissions place the reward in a lower band that accurate assessments on other dimensions cannot offset (Appendix~\ref{app:judge_training_objective}). Section~\ref{sec:exp_judge} evaluates these stages through both scoring error and CSA.

\paragraph{Returning the updated judge to the workflow.}
After evaluation, a candidate $J^{(k+1)}$ can replace the in-loop judge in Section~\ref{sec:query_rewrite}: its scores enter Equation~\ref{eq:score}, and its reasons guide prompt repair. Subsequent runs then produce records for another offline update. Collection, post-training, evaluation, and redeployment define the proposed self-evolution cycle. Training annotation, in-loop verification, and external CSA evaluation serve distinct roles; external test verdicts are not fed back into generation. The reported judge configurations evaluate the post-training stages, while repeated collection--training cycles remain a proposed extension.

% ============================================================
%  4  EXPERIMENTS
% ============================================================

\section{Experiments}
\label{sec:experiments}

We organize our experiments around three questions: (i) how ImageRSI performs end to end compared with generative models and agent systems (\S\ref{sec:exp_main}); (ii) how localized repair and experience reuse each contribute to its gains (\S\ref{sec:exp_repair} and \S\ref{sec:exp_experience}); and (iii) whether post-training the judge on the system's accumulated judgment records improves end-to-end performance (\S\ref{sec:exp_rsi}).

% ---------------------------------------------------------------------
\subsection{Main Results}
\label{sec:exp_main}

\subsubsection{Experimental Setup}

\paragraph{Benchmarks.}
\label{sec:benchmarks}
MindBench~\citep{yang2026mindbench} contains 500 image-generation requests requiring world knowledge or reasoning across ten categories; three categories (Life, GU, and Math) involve image-to-image (I2I) tasks with an input image. ImageReq-Bench, constructed in this work, contains 961 requests across four reporting categories: Illustration, Scene, Infographic, and Design \& Art. Each request is paired with an evaluation checklist, yielding 8,165 check items in total. We use a category-stratified split into an experience set (530 requests), used to construct the experience library, and a disjoint test set (431 requests) for evaluation. Table~\ref{tab:imagereq} reports results on the test set.

\paragraph{Metric.}
We adopt Checklist Strict Accuracy (CSA): a request counts as passed only if the delivered image satisfies every item on its checklist. Requests without a generated image or with an evaluator error count as failures, so the denominator includes every request in the evaluation set. The external evaluator is Gemini 3.1 Pro; it is independent of the system's internal judge, and its verdict on a request is never fed back into the generation of that request.

\paragraph{Implementation details.}
The generator is always frozen: Qwen-Image-2.0 for the main MindBench results and the judge experiments (\S\ref{sec:exp_rsi}), and Wan2.7-Image elsewhere. The internal judge scores each image along five dimensions: Alignment, Faithfulness, Quality, Aesthetic, and Physics (\S\ref{sec:verification}). An image is accepted if its weighted score is at least 4.5 and no veto is triggered; otherwise, the system repairs the localized defects, using zero, one, or two repair rounds per request. For each baseline, we report the higher of our reproduction and the value claimed in its original paper.

\subsubsection{Results on MindBench}

As shown in Table~\ref{tab:mindbench}, ImageRSI achieves an overall CSA of 48.2\%, outperforming all baselines. It improves over direct generation with its own generator, Qwen-Image-2.0 (22.4\%), by 25.8 percentage points (pp), and over the strongest agent baseline, GEMS \cite{he2026gems}(42.4\%), by 5.8\,pp; with Wan2.7-Image, ImageRSI reaches 47.4\%. Per category, ImageRSI ranks first on IP (30\%) and leads most clearly on Math (80\% vs.\ 46\% for Nano Banana Pro).

Replacing the generator with Nano Banana Pro raises the overall CSA of ImageRSI to 52.6\% (263/500), 4.4\,pp higher than the Qwen-Image-2.0 version; the difference comes mainly from SE (70\% vs.\ 48\%) and SL (72\% vs.\ 56\%).

\begin{table*}[t]
\centering
\small
\setlength{\tabcolsep}{4pt}
\begin{tabular}{lccccccccccc}
\toprule
\textbf{Method} & \textbf{SE} & \textbf{Weather} & \textbf{MC} & \textbf{IP} & \textbf{WK} & \textbf{SL} & \textbf{Poem} & \textbf{Life} & \textbf{GU} & \textbf{Math} & \textbf{Overall} \\
\midrule
\multicolumn{12}{l}{\textit{Generative models}} \\
SD-3.5-M & 0.02 & 0.00 & 0.00 & 0.00 & 0.02 & 0.00 & 0.00 & -- & -- & -- & 0.00 \\
Bagel~\citep{deng2025emerging} & 0.02 & 0.00 & 0.00 & 0.00 & 0.00 & 0.02 & 0.02 & 0.02 & 0.00 & 0.08 & 0.02 \\
Qwen-Image~\citep{wu2025qwen} & 0.08 & 0.00 & 0.04 & 0.00 & 0.00 & 0.04 & 0.00 & 0.04 & 0.00 & 0.00 & 0.02 \\
Echo-4o~\citep{ye2025echo} & 0.04 & 0.00 & 0.00 & 0.00 & 0.00 & 0.02 & 0.06 & 0.02 & 0.02 & 0.02 & 0.02 \\
DraCo~\citep{jiang2026draco} & 0.02 & 0.00 & 0.02 & 0.00 & 0.00 & 0.02 & 0.02 & 0.04 & 0.02 & 0.06 & 0.02 \\
Wan2.7-Image & 0.00 & 0.02 & 0.24 & 0.06 & 0.00 & 0.20 & 0.06 & 0.06 & 0.08 & 0.12 & 0.084 \\
GPT-Image-1 & 0.32 & 0.06 & 0.22 & 0.02 & 0.16 & 0.32 & 0.10 & 0.24 & 0.10 & 0.12 & 0.17 \\
Nano Banana & 0.30 & 0.10 & 0.12 & 0.00 & 0.30 & 0.32 & 0.36 & 0.20 & 0.04 & 0.08 & 0.18 \\
FLUX 2 Pro & 0.38 & 0.12 & 0.08 & 0.00 & 0.20 & 0.44 & 0.64 & 0.18 & 0.04 & 0.02 & 0.21 \\
GPT-Image-1.5 & 0.36 & 0.18 & 0.22 & 0.04 & 0.30 & 0.34 & 0.08 & \textbf{0.34} & 0.10 & 0.02 & 0.198 \\
FLUX 2 Max & 0.44 & 0.12 & 0.10 & 0.04 & 0.38 & 0.40 & 0.50 & 0.20 & 0.02 & 0.06 & 0.23 \\
Qwen-Image-2.0~\citep{zhao2026qwen} & 0.19 & 0.24 & 0.23 & 0.04 & 0.12 & 0.42 & 0.58 & 0.12 & 0.02 & 0.28 & 0.224 \\
Seedream 4.5 & 0.16 & 0.18 & 0.30 & 0.06 & 0.32 & 0.36 & 0.58 & 0.18 & \underline{0.30} & 0.24 & 0.268 \\
Nano Banana Pro & 0.50 & \underline{0.36} & 0.40 & 0.16 & 0.56 & 0.62 & 0.68 & 0.30 & 0.16 & 0.46 & 0.420 \\
\midrule
\multicolumn{12}{l}{\textit{Agent systems}} \\
Mind-Brush & 0.54 & 0.16 & 0.62 & 0.18 & 0.40 & 0.26 & 0.54 & 0.10 & 0.16 & 0.14 & 0.31 \\
GenSearcher-style~\citep{feng2026gen} & 0.42 & 0.34 & \underline{0.70} & 0.26 & \textbf{0.66} & \underline{0.64} & 0.08 & \underline{0.32} & \underline{0.30} & 0.44 & 0.416 \\
Qwen-Image-Agent & \underline{0.60} & 0.28 & \underline{0.70} & 0.16 & 0.28 & 0.58 & \textbf{0.82} & 0.24 & 0.20 & 0.34 & 0.42 \\
GEMS & 0.38 & \textbf{0.48} & 0.56 & 0.16 & \underline{0.60} & 0.44 & 0.66 & 0.18 & \textbf{0.38} & 0.40 & 0.424 \\
\rowcolor{oursrow}
\textbf{ImageRSI (Wan2.7-Image)} & 0.42 & 0.30 & \textbf{0.72} & \underline{0.28} & 0.46 & 0.56 & 0.72 & 0.28 & 0.22 & 0.78 & 0.474 \\
\rowcolor{oursrow}
\textbf{ImageRSI (Qwen-Image-2.0)} & 0.48 & 0.30 & 0.62 & \textbf{0.30} & 0.50 & 0.56 & 0.74 & 0.30 & 0.22 & \underline{0.80} & \underline{0.482} \\
\rowcolor{oursrow}
\textbf{ImageRSI (Nano Banana Pro)} & \textbf{0.70} & 0.24 & 0.62 & 0.26 & 0.58 & \textbf{0.72} & \underline{0.76} & 0.30 & 0.26 & \textbf{0.82} & \textbf{0.526} \\
\bottomrule
\end{tabular}
\caption{Per-category CSA on MindBench. Best results are in bold and second-best ones are underlined; ImageRSI rows are shaded. SE: special events; MC: new movie characters; IP: popular IPs; WK: world knowledge; SL: science and logic; GU: map-to-view.}
\label{tab:mindbench}
\end{table*}

\subsubsection{Results on ImageReq-Bench}

As shown in Table~\ref{tab:imagereq}, ImageRSI achieves the highest overall CSA, 65.66\%. It improves over direct generation with Wan2.7-Image (38.28\%) by 27.38\,pp; over the strongest generative model, Nano Banana (56.38\%), and the strongest agent baseline, GEMS (55.68\%), by 9.28 and 9.98\,pp, respectively; and over MemoGen R2 (47.56\%) and SCOPE~\citep{ren2026scope} (46.87\%) by 18.10 and 18.79\,pp. Per category, ImageRSI ranks first on Illustration, Scene, and Infographic. On Illustration and Scene, which together account for 75.2\% of the test set, it leads the runner-up, GEMS, by 9.17 and 11.58\,pp.

\begin{table}[t]
\centering
\small
\setlength{\tabcolsep}{3.2pt}
\begin{tabular}{lrrrrr}
\toprule
\textbf{Method} & \textbf{Illus.} & \textbf{Scene} & \textbf{Info.} & \textbf{D\&A} & \textbf{All} \\
\midrule
\multicolumn{6}{l}{\textit{Generative models}} \\
Qwen-Image-2.0 & 38.43 & 35.79 & 35.38 & 16.67 & 35.27 \\
Wan2.7-Image & 42.36 & 30.53 & 35.38 & 38.10 & 38.28 \\
Seedream 4.5 & 49.34 & 35.79 & 40.00 & 52.38 & 45.24 \\
Nano Banana & 57.64 & 50.53 & 50.77 & \textbf{71.43} & \underline{56.38} \\
\midrule
\multicolumn{6}{l}{\textit{Agent systems}} \\
Mind-Brush & 39.74 & 35.79 & 32.31 & 28.57 & 36.66 \\
Qwen-Image-Agent & 46.29 & 33.68 & 36.92 & 40.48 & 41.53 \\
GenSearcher-style & 49.78 & 40.00 & 35.38 & 45.24 & 45.01 \\
SCOPE & 50.22 & 38.95 & 43.08 & 52.38 & 46.87 \\
MemoGen R2 & 54.15 & 38.95 & 38.46 & 45.24 & 47.56 \\
GEMS & \underline{59.83} & \underline{52.63} & \underline{52.31} & 45.24 & 55.68 \\
\rowcolor{oursrow}
\textbf{ImageRSI} & \textbf{69.00} & \textbf{64.21} & \textbf{53.85} & \underline{69.05} & \textbf{65.66} \\
\bottomrule
\end{tabular}
\caption{CSA (\%) on the ImageReq-Bench test set by category. Best results are in bold and second-best ones are underlined. Illus.: Illustration; Info.: Infographic; D\&A: Design \& Art.}
\label{tab:imagereq}
\end{table}

% ---------------------------------------------------------------------
\subsection{Effectiveness of Verification and Repair}
\label{sec:exp_repair}

We first test whether the gain from verification and repair comes only from generating one additional image, by comparing second-round strategies under the same generation budget.

\paragraph{Experimental settings.}
The repair and experience-reuse ablations (\S\ref{sec:exp_repair} and \S\ref{sec:exp_experience}) use the test set defined in \S\ref{sec:benchmarks}, with Wan2.7-Image for generation and Gemini 3.1 Pro for evaluation.

\paragraph{Compared strategies.}
All strategies share the same first-round image. If the judge accepts it, the image is delivered directly; otherwise, one of the following strategies generates a second image. Each request therefore receives at most two images, and all strategies have the same generation budget.
(i) \textbf{Best-of-2 resampling:} regenerate with the same prompt and let the judge select the better image;
(ii) \textbf{Generic-feedback repair:} ask the rewriter to revise the original prompt without specifying the defects;
(iii) \textbf{Localized repair (ImageRSI):} specify the localized defects in the rewrite instruction and lock the parsed requirements (\S\ref{sec:req_parse}) so that satisfied requirements are preserved.
We also ablate requirement locking by removing it from localized repair, keeping all else unchanged.

\begin{figure*}[t]
\centering
\includegraphics[width=\textwidth]{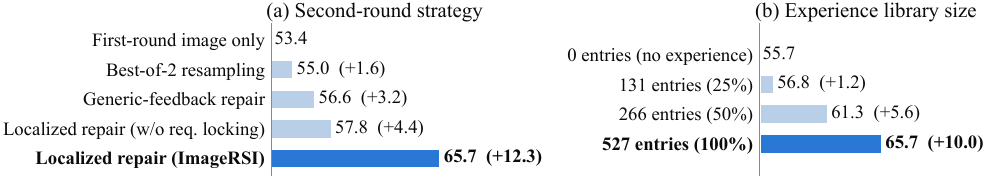}
\caption{Ablation studies of repair strategies and experience library size on the held-out test set. (a)~Second-round strategies with a shared first-round image and at most two images per request; bars show the gain over the first-round image only (pp), and labels give CSA (\%) with the gain. (b)~CSA with nested experience libraries of 0, 131, 266, and 527 entries.}
\label{fig:ablation}
\end{figure*}

\paragraph{Results.}
As shown in Figure~\ref{fig:ablation}a, localized repair raises CSA from 53.4\% to 65.7\% (+12.3\,pp), 10.7\,pp above Best-of-2 resampling, which gains only 1.6\,pp. The gain therefore does not come from the additional generation alone. Generic-feedback repair gains only 3.2\,pp, indicating that signaling failure alone is of limited help. Without requirement locking, CSA drops to 57.8\%, 7.9\,pp below full localized repair, suggesting that locking contributes substantially to the gain.

\begin{figure*}[t]
\centering
\includegraphics[width=\textwidth]{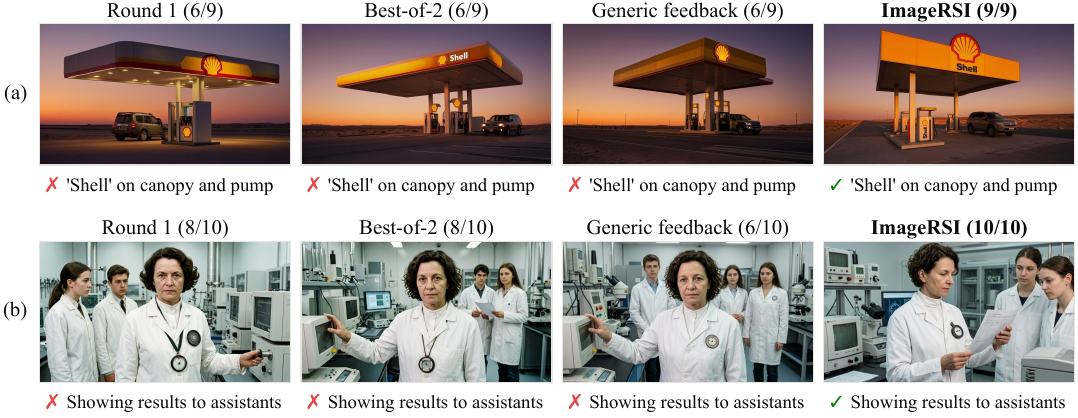}
\caption{Verification-and-repair cases from the test set: (a) case 1006; (b) case 0656. Each row shows the first-round image and the second-round image of each strategy; parentheses give the passed check items, and marks give the external verdict on the key item failed in the first round. In (b), the judge ultimately delivers the first-round image under Best-of-2 resampling and generic-feedback repair.}
\label{fig:repair_case}
\end{figure*}

Figure~\ref{fig:repair_case} presents two cases. Case~(a) requires a desert roadside gas station to display a complete, legible Shell emblem, comprising the scallop-shell graphic and the brand name ``Shell,'' on both the canopy front and the fuel-pump casing. In the first-round image, both emblems show only the graphic without the text, failing the three text-related check items (6/9); the judge correctly localizes the missing brand name. Case~(b) requires the main character to present measurement results to two assistants in a radiation laboratory, with detailed constraints on her appearance and clothing and on how the two assistants differ. In the first-round image, she faces forward while operating an instrument, without interacting with the assistants, who are also not standing behind the instruments; two items fail (8/10), and the judge identifies the missing interaction with the assistants. In both cases, the second images from Best-of-2 resampling and generic-feedback repair fail to fix these errors; in case~(b), the generic-feedback image even contains three assistants and loses two previously satisfied requirements (6/10). Localized repair instead states the defect in the rewritten prompt and excludes it in the negative prompt (``graphic only, no text''; ``facing forward and operating an instrument, no interaction''), after which all items pass in both cases.

% ---------------------------------------------------------------------
\subsection{Experience Reuse}
\label{sec:exp_experience}

\begin{table}[t]
\centering
\small
\setlength{\tabcolsep}{6pt}
\begin{tabular}{lcc}
\toprule
\textbf{Internal judge} & \textbf{MAE} $\downarrow$ & \textbf{MindBench CSA (\%)} \\
\midrule
\multicolumn{3}{l}{\textit{Frozen closed-source models}} \\
Qwen3.6-Plus & -- & 24.2 \\
Qwen3.7-Plus & -- & 43.4 \\
GPT-6-Astra & -- & 44.2 \\
\midrule
\multicolumn{3}{l}{\textit{Qwen3.8-27B with judge training}} \\
Base (untrained) & 0.598 & 38.6 \\
SFT$_{30\%}$ & 0.596 & 39.8 \\
SFT$_{60\%}$ & 0.582 & 40.2 \\
SFT$_{100\%}$ & \underline{0.453} & \underline{44.8} \\
\rowcolor{oursrow}
\textbf{GRPO}$_{\mathbf{100\%}}$ & \textbf{0.403} & \textbf{48.2} \\
\bottomrule
\end{tabular}
\caption{End-to-end MindBench results with different internal judges. Best values are in bold and second-best values are underlined. MAE is computed against teacher scores on the same 90 images (--: not reported); end-to-end runs use the Qwen-Image-2.0 generator, at most two repair rounds, and a frozen experience library.}
\label{tab:judge}
\end{table}

Beyond repairing a single request, ImageRSI reuses experience across requests. We examine whether experience accumulated from past trajectories improves performance on new requests, and whether the gain grows with library size.

\paragraph{Experience construction and use.}
Taking the final prompt delivered to the generator as the trajectory and the external evaluator's per-item verdicts as supervision, Qwen3.8-Max distills one candidate experience per request in the experience set. Each candidate specifies the applicable request type (with proper nouns, numbers, and text content removed), at most three sentences of prompt-writing advice, and the check items most prone to failure for such requests. We retain 527 experiences after applying the admission protocol in Appendix~\ref{app:admission}. At inference time, the system retrieves the three most similar experiences by character-bigram TF-IDF similarity and injects them into prompt rewriting in both the first round and the repair round; all other settings follow localized repair in \S\ref{sec:exp_repair}.

\paragraph{Library size.}
We arrange the experience-set requests in a fixed random order and take successive prefixes to build four nested libraries with 0, 131, 266, and 527 experiences, corresponding to roughly 0\%, 25\%, 50\%, and 100\% of the full library. The libraries differ only in size.

\paragraph{Results.}
As shown in Figure~\ref{fig:ablation}b, CSA increases monotonically with library size: 55.7\%, 56.8\%, 61.3\%, and 65.7\%. The full library improves CSA by 10.0\,pp over using no experience and exceeds the 25\% and 50\% libraries by 8.8 and 4.4\,pp, respectively. The benefit of experience reuse thus grows with the size of the library.

% ---------------------------------------------------------------------
\subsection{Recursive Self-Improvement}
\label{sec:exp_rsi}
\label{sec:exp_judge}

ImageRSI improves itself at two levels. Within a task, it revises its output round by round based on judge feedback; across tasks, it distills trajectories into experience and updates the judge with accumulated judgment records. This section evaluates the judge update (\S\ref{sec:judge_evolution}).

\paragraph{Setup.}
We train Qwen3.8-27B with supervised fine-tuning (SFT) on 30\%, 60\%, and 100\% of the SFT training set, obtaining SFT$_{30\%}$, SFT$_{60\%}$, and SFT$_{100\%}$. We then train GRPO$_{100\%}$ with GRPO initialized from SFT$_{100\%}$ on a separate GRPO training set whose requests do not overlap with the SFT set, keeping the SFT output format. The GRPO reward combines pointwise assessment with within-group ranking (Appendix~\ref{app:judge_training_objective}). The pointwise term scores both the five-dimensional scores, through the weighted scoring error and per-dimension agreement with the target label at the pass boundary (score $\geq 4$), and the rationales, by checking whether each criticism is supported by the image and the request. Malformed outputs receive the lowest reward; accepting an image whose target alignment or faithfulness score is at most 2, or making a critical fabrication or omission, moves the reward into a separate lower band that accurate assessments on other dimensions cannot offset. We report the mean absolute error (MAE) against teacher-model reference scores and the end-to-end CSA when each judge is plugged into the full pipeline. For reference, we also report results with off-the-shelf multimodal LLMs used as the internal judge without training.

\paragraph{Results.}
As shown in Table~\ref{tab:judge}, as the training data grows from 30\% to 100\%, the MAE of the SFT judges decreases from 0.596 to 0.453 (base model: 0.598), and CSA increases from 39.8\% to 44.8\% (base model: 38.6\%). GRPO further reduces MAE to 0.403 and raises CSA to 48.2\%, the best on both metrics, improving over SFT$_{100\%}$ by 0.050 and 3.4\,pp, respectively. Off-the-shelf judges yield CSA between 24.2\% (Qwen3.6-Plus) and 44.2\% (GPT-6-Astra), showing that the choice of judge markedly affects system performance. Base model without updates (38.6\%) falls below Qwen3.7-Plus and GPT-6-Astra, whereas after SFT and GRPO (48.2\%) it surpasses all off-the-shelf judges, exceeding GPT-6-Astra by 4.0\,pp. Training on the system's own accumulated judgment records thus enables the 27B judge to outperform every off-the-shelf judge we evaluated.

\section{Conclusion}
\label{sec:conclusion}

ImageRSI couples five-dimensional verification and targeted repair with a validated cross-task experience library and offline judger self-evolution under a frozen image generator. It reaches 52.6\% CSA on MindBench with Nano Banana Pro and 65.66\% on ImageReq-Bench. Under a matched generation budget, localized repair with requirement locking outperforms resampling and generic feedback, and CSA increases with the size of the experience library; post-training the judge on accumulated judgment records further improves end-to-end CSA. The gains from repeated collection--training cycles remain to be evaluated.

\section*{Limitations}

We evaluate ImageRSI with several base image generators. Our focus is on improving generation quality for a given generator through verification, repair, and cross-task experience reuse. A related application arises when the base generator is replaced. Here, existing experience could be reused to reduce the cost of accumulating it again. For example, one could study which experiences transfer directly and which need to be adapted to the characteristics of the new generator. Such optimizations could be added to the existing pipeline as an optional extension of the experience library, and they are independent of the contributions of our method.

\bibliography{custom}

\clearpage
\appendix
% ============================================================
%  APPENDIX (included from acl_latex.tex after \appendix)
% ============================================================
% Box styles for Appendix C. If the preamble already defines them (tcolorbox), those are used;
% otherwise the plain-LaTeX fallback below is used, so this file compiles with any preamble.
\makeatletter
\@ifundefined{pfield}{\newcommand{\pfield}[1]{\par\noindent\textbf{#1:}\ }}{}
\@ifundefined{promptbox}{%
  \definecolor{pframe}{HTML}{4A6FA5}\definecolor{pback}{HTML}{F7F9FC}\definecolor{ptitle}{HTML}{DCE6F2}%
  \definecolor{eframe}{HTML}{5B8C5A}\definecolor{eback}{HTML}{F6FAF5}\definecolor{etitle}{HTML}{E1EEDF}%
  \newsavebox{\apx@box}%
  \newenvironment{apx@frame}[4]{%
    \def\apx@f{#1}\def\apx@b{#2}\par\smallskip\noindent
    \begin{lrbox}{\apx@box}\begin{minipage}{\dimexpr\columnwidth-2\fboxsep-2\fboxrule\relax}\small
    {\setlength{\fboxsep}{2pt}\colorbox{#3}{\parbox{\dimexpr\linewidth-4pt\relax}{\bfseries #4}}}\par\smallskip}%
   {\end{minipage}\end{lrbox}\fcolorbox{\apx@f}{\apx@b}{\usebox{\apx@box}}\par\smallskip}%
  \newenvironment{promptbox}[1]{\begin{apx@frame}{pframe}{pback}{ptitle}{#1}}{\end{apx@frame}}%
  \newenvironment{expbox}[1]{\begin{apx@frame}{eframe}{eback}{etitle}{#1}}{\end{apx@frame}}%
}{}
\makeatother

\noindent\textbf{Appendix organization.}
Appendix~\ref{app:implementation} gives implementation details; Appendix~\ref{app:data_evaluation} describes the datasets and the CSA metric; Appendix~\ref{app:experience_prompts} presents experience records and prompt interfaces; Appendix~\ref{app:training_reproducibility} describes judge training; Appendix~\ref{app:exp_details} gives additional experimental details; and Appendix~\ref{app:supp_figures} contains a supplementary figure.

\section{Implementation Details}
\label{app:implementation}

\subsection{Generation and Delivery}
\label{app:verification}
\label{app:generation_delivery}
Retrieval is limited to four passes per request. Each round generates one image, and a request receives an initial round followed by at most two repair rounds; the direct-generation baseline uses a single round. In each round, the workflow selects experience, builds the prompt from the fixed requirements and earlier feedback, generates an image, and scores it. The first accepted image is returned. If no round is accepted within the budget, the highest-scoring judged image is returned as a fallback; it is not counted as accepted. Failed generation or judging calls are recorded in the trajectory.

\subsection{Scoring and Acceptance}
\label{app:scoring_acceptance}
Each dimension is scored on an integer scale from 1 to 5. For T2I requests, the base weights of Alignment, Faithfulness, Quality, Aesthetic, and Physics are $0.40$, $0.20$, $0.25$, $0.20$, and $0.15$. Following Equation~\ref{eq:score}, the weight of a low-confidence dimension is multiplied by $0.7$, and the weights are then normalized over the active dimensions. I2I requests additionally activate a reference-consistency check.

An image is accepted if its aggregate score is at least $4.5$ and no veto applies. A Quality score of at most 3 vetoes acceptance; so does an Alignment score of at most 3, unless the requirement parser reports low confidence.

\subsection{Constraint-Aware Verification}
\label{app:constraint_verification}
In this optional variant, an extractor splits the request into atomic requirements, each linked to its source phrase in the request. The verifier labels every requirement \texttt{pass}, \texttt{fail}, or \texttt{unknown} and cites visible evidence; counts that are occluded or relations that cannot be seen remain \texttt{unknown}. Early acceptance requires every requirement to pass. The repair prompt keeps the clauses of satisfied requirements and rewrites only those for failed or unknown ones; satisfied requirements are rechecked after repair.

\subsection{Experience Admission}
\label{app:admission}
A candidate needs at least two independent source trajectories. Each validation trial applies only the candidate's action, with the seed and all upstream inputs shared between the two arms, and the image pair is scored blind. Calibration and holdout use separate splits. Calibration requires at least three trials, at least two positive differences $\Delta_e(q)$, no negative difference, and a mean difference of at least $0.05$. Holdout requires at least two trials, at least one positive difference, no negative difference, and a nonnegative mean. A candidate becomes \texttt{verified} after calibration and \texttt{active} after holdout; any hard regression rejects it.

\section{Datasets and Evaluation}
\label{app:data_evaluation}

\subsection{ImageReq-Bench}
\label{app:datasets}

ImageReq-Bench is built around hard cases in real-world text-to-image requests, i.e., requests whose requirements current generators often fail to satisfy jointly. Benchmark scope and the experience/test split are described in \S\ref{sec:benchmarks}. During curation, we deduplicated candidate requests and removed or replaced identified personal information. Each request is linked to its source records. All methods and judges are evaluated on the same fixed list of test case IDs.

We consolidate the 12 content categories into four: Illustration (511 requests), Scene (213), Infographic (144), and Design \& Art (93). These categories are organizational labels rather than a validated difficulty taxonomy.

Language or vision--language models were used to draft and refine a checklist for each request, removing unsupported requirements, merging duplicate items, and expressing the remaining requirements as visually checkable statements. Checklists contain 2 to 18 items per request, and all methods are scored against the same checklist version. The benchmark assesses whether multiple requirements within a request are jointly satisfied and uses individual checks to identify unmet requirements, providing a common basis for evaluating image generation, targeted repair, and cross-task experience reuse.

The three MindBench I2I categories (Life, GU, and Math) comprise 150 requests.

\subsection{Checklist Strict Accuracy}
\label{app:csa_protocol}

Let $b_{ij}$ be the external evaluator's Boolean verdict on checklist item $j$ of request $i$. Request $i$ passes ($y_i=1$) only if a delivered image exists, the evaluator returns a valid verdict for every item, and all $b_{ij}$ are true; otherwise $y_i=0$. For an evaluation set of $N$ requests,
\begin{equation}
\label{eq:appendix_csa}
\operatorname{CSA}=\frac{1}{N}\sum_{i=1}^{N}y_i.
\end{equation}
Malformed evaluator responses are never parsed as passes.

The external evaluator, Gemini 3.1 Pro, is separate from the internal judge. Its verdicts on a request are never fed back into the generation of that request; experience is distilled only from verdicts on the experience set (\S\ref{sec:exp_experience}).

\section{Experience Records and Prompt Interfaces}
\label{app:experience_prompts}

\subsection{Experience Record Structure}
\label{app:experience_records}

Table~\ref{tab:experience_fields} lists the fields of an experience record. During selection, the model first sees only the Domain, Task Type, Generation Type, and Failure Pattern of each entry, and then loads the selected entries in full.

\begin{table*}[t]
\centering
\small
\begin{tabular}{lp{0.65\textwidth}}
\toprule
Field & Content \\
\midrule
Domain, Task Type, Generation Type & Subject area and input modality, used to narrow retrieval. \\
Failure Pattern, Trigger & The observed failure or request condition under which the strategy applies. \\
Negative Trigger & Conditions under which the strategy should not be used. \\
Action, protected constraints & The prompt change and the requirements it must keep. \\
Effect, Known Side Effects & Observed outcomes in trials and possible regressions. \\
Source, trial records, state & Supporting trajectories, paired validation trials, and admission state. \\
\bottomrule
\end{tabular}
\caption{Fields of an experience record.}
\label{tab:experience_fields}
\end{table*}

\subsection{Example Entries}
\label{app:operator_examples}

Two entries from the experience library are shown below, using the fields of Table~\ref{tab:experience_fields}.

\begin{expbox}{\texttt{preserve-identity-anchors}}
\pfield{Trigger} A named subject is generalized, confused with another person, or hard to tell apart from other subjects.
\pfield{Negative Trigger} Anonymous subjects, intentionally hidden faces, or images whose identities are already correct.
\pfield{Action} Add a separate identity description for each subject, using only attributes supported by the request or by retrieved references.
\pfield{Protected} The requested pose, clothing, scene, and style.
\pfield{Known Side Effects} Less compositional freedom; extra text that may itself be rendered incorrectly.
\end{expbox}

\begin{expbox}{\texttt{explicit-spatial-anchoring}}
\pfield{Trigger} Object placement or relation binding is ambiguous.
\pfield{Negative Trigger} Requests that leave the arrangement open, or failures that concern only style or lighting.
\pfield{Action} Name each target and reference object, state the direction or containment explicitly, and list discrete elements when a count is required; for example, \emph{place A nearby} is replaced with an explicit reference object and spatial relation.
\pfield{Protected} Subject identity, exact text, counts, and other appearance constraints.
\pfield{Known Side Effects} Longer prompts and less layout freedom.
\end{expbox}

\subsection{Prompt Interfaces}
\label{app:prompt_contracts}

The boxes below summarize the input and output of each prompt. Prompts~1--5 belong to the task-level workflow (\S\ref{sec:query_rewrite}), Prompt~6 to experience admission (\S\ref{sec:experience}), and Prompt~7 to the optional constraint-aware variant (Appendix~\ref{app:constraint_verification}); Prompts~8 and~9 are specific to the controlled experiments in \S\ref{sec:exp_repair} and \S\ref{sec:exp_experience}. Angle brackets denote placeholders.

\begin{promptbox}{Prompt 1: Requirement Parsing}
\pfield{Input} Request $q_t$, permitted evidence, and optional input image $i_t$.
\pfield{Output} Requirement set $C_t$ covering entities, attributes, counts, spatial relations, exact text, layout, and I2I preservation or exclusion conditions, together with their bindings.
\pfield{Rule} $C_t$ stays fixed across repair rounds; a revision must satisfy an unmet requirement rather than drop it.
\end{promptbox}

\begin{promptbox}{Prompt 2: Experience Selection}
\pfield{Input} Request, requirements, the Domain, Task Type, Generation Type, and Failure Pattern of each entry, and, after a failed round, the observed defects.
\pfield{Output} Names of applicable entries, which are then loaded in full; entries whose Negative Trigger matches are excluded.
\end{promptbox}

\begin{promptbox}{Prompt 3: Prompt Rewriting}
\pfield{Input} Request $q_t$, fixed requirements $C_t$, evidence $E_t$, selected experience $e_{t,r}$, and feedback from earlier rounds $f_{<r}$.
\pfield{Output} Generation prompt $p_{t,r}$ that keeps entities, exact text, counts, and relations.
\pfield{Localized repair} State the localized defects in the rewritten prompt and exclude them in the negative prompt.
\pfield{Example} Case 1006 (Figure~\ref{fig:repair_case}a). Defect: both Shell emblems show only the scallop-shell graphic, without the brand name. Negative prompt: ``graphic only, no text''.
\end{promptbox}

\begin{promptbox}{Prompt 4: Dimension Scoring}
\pfield{Input} Request, parsed requirements, generated image, and references if any.
\pfield{Output} For each active dimension, a score from 1 to 5, defect evidence, and a confidence value.
\pfield{Rule} Scores rely only on the request, the requirements, and the image evidence; the generation prompt and selected experience are not shown.
\pfield{Trained-judge target} Five lines in a fixed order (schematic):
{\scriptsize
\begin{verbatim}
Alignment:    <1-5>  <reason grounded in the image>
Faithfulness: <1-5>  <reason grounded in the image>
Quality:      <1-5>  <reason grounded in the image>
Physics:      <1-5>  <reason grounded in the image>
Aesthetic:    <1-5>  <reason grounded in the image>
\end{verbatim}
}
\end{promptbox}

\begin{promptbox}{Prompt 5: Attribution}
\pfield{Input} Dimension scores, image, generation prompt, and selected experience.
\pfield{Output} Repair suggestions; the scores are not changed.
\end{promptbox}

\begin{promptbox}{Prompt 6: Paired Experience Review}
\pfield{Input} Request and a blinded image pair generated with and without candidate $e$, sharing the seed and upstream inputs.
\pfield{Output} A score for each image; their difference gives $\Delta_e(q)$ in Equation~\ref{eq:experience_delta}.
\end{promptbox}

\begin{promptbox}{Prompt 7: Constraint Checking (optional variant)}
\pfield{Input} Atomic requirements of the original request and the current image.
\pfield{Output} Each requirement ID exactly once, with a status and visible evidence:
{\scriptsize
\begin{verbatim}
{"constraints": [{
  "id": "<requirement ID>",
  "status": "pass | fail | unknown",
  "evidence": "<visible evidence>"
}]}
\end{verbatim}
}
\pfield{Rule} A visible contradiction counts as \texttt{fail}; when the evidence is insufficient, such as an occluded count, the item remains \texttt{unknown}.
\end{promptbox}

\begin{promptbox}{Prompt 8: Generic-Feedback Repair (ablation, \S\ref{sec:exp_repair})}
\pfield{Instruction} The image did not pass; improve the prompt against the original request.
\pfield{Not provided} Any information about the defects.
\end{promptbox}

\begin{promptbox}{Prompt 9: Case-Level Experience Distillation (\S\ref{sec:exp_experience})}
\pfield{Model} Qwen3.8-Max.
\pfield{Input} Final prompt delivered to the generator and the external evaluator's per-item verdicts, for experience-set requests only.
\pfield{Output} (i) The applicable request type, with proper nouns, numbers, and text content removed; (ii) at most three sentences of prompt-writing advice; (iii) the check items most prone to failure for such requests.
\pfield{Use} The three most similar entries by character-bigram TF-IDF are injected into prompt rewriting in both the first round and the repair round.
\end{promptbox}

\section{Judge Training}
\label{app:training_reproducibility}

\subsection{Training Data}
\label{app:supervision_data}

\paragraph{Training requests.}
We assemble approximately 2,000 requests for judge training, combining 530 requests from the experience split of ImageReq-Bench with approximately 1,500 additional requests. The latter are constructed using WISE's taxonomy and prompt design as guidance, together with MindBench's knowledge- and reasoning-driven task categories~\citep{niu2025wise,yang2026mindbench}. WISE provides sampled coverage of world knowledge, while MindBench contains 50 evaluation examples per category. Although both offer useful evaluation resources, neither directly supplies the five-dimensional score-and-reason supervision required by our judge. We therefore expand the request pool along their task categories and pair selected images from generation and repair trajectories with dimension-specific scores and evidence-grounded explanations.

\paragraph{Inputs and targets.}
Each training input contains the request, the parsed requirements, permitted evidence, any task input image, and the generated image. The target consists of five lines, one per dimension in the order Alignment, Faithfulness, Quality, Physics, and Aesthetic; each line gives an integer score from 1 to 5 and a reason grounded in the image. Images that violate the request are kept, since their targets describe the violation. Training covers only scores and reasons. The confidence values used in Equation~\ref{eq:score} are set to constants by an inference adapter when a trained judge is deployed.

\subsection{Training Objective}
\label{app:judge_training_objective}

Let $z$ be the judge input and $y=(y_1,\ldots,y_L)$ the target response. SFT minimizes
\begin{equation}
\label{eq:judge_response_loss}
\mathcal{L}_{\mathrm{SFT}}(\theta)
=-\frac{1}{L}\sum_{\ell=1}^{L}
\log p_{\theta}(y_{\ell}\mid z,y_{<\ell}),
\end{equation}
with prompt and image tokens masked from the loss. GRPO then starts from the SFT$_{100\%}$ checkpoint.

\paragraph{GRPO reward.}
The GRPO reward combines pointwise assessment with within-group ranking. Pointwise assessment evaluates five-dimensional score accuracy and rationale correctness, checking whether the judge's criticisms are supported by the image and task requirements and identifying overly harsh assessments or other misjudgments. Malformed outputs receive the lowest reward, while severe false accepts and critical fabrications or omissions enter a separate lower reward band that cannot be offset by other dimensions. Within each band, a small ranking adjustment is added only when two reversed-order reviews yield consistent, evidence-supported rankings with sufficient confidence. Score aggregation and acceptance rules are fixed and not learned.

\section{Experimental Details}
\label{app:exp_details}

\subsection{Baselines on MindBench}
\label{app:baselines}

Table~\ref{tab:mindbench} combines values reported in the original papers with our reproductions; for each baseline, we report the higher of the two (\S\ref{sec:exp_main}). GEMS and GenSearcher-style are reproduced locally with the Qwen-Image-2.0 generator and a Gemini 3.1 Pro backbone; image retrieval is unavailable in our GenSearcher-style reproduction. Seedream 4.5 is evaluated with direct generation. Wan2.7-Image is run locally; one request remained unevaluated and is counted as a failure, so its CSA is a lower bound. For ImageRSI (Nano Banana Pro), 12 requests produced no image and are counted as failures.

\subsection{Repair Strategies}
\label{app:repair_details}

In generic-feedback repair, the rewriter is told only that the image did not pass and is asked to improve the prompt against the original request; no defect information is given. Table~\ref{tab:repair_ablation} lists the values plotted in Figure~\ref{fig:ablation}a. The two cases in Figure~\ref{fig:repair_case} use seeds 12345 (case 1006) and 22345 (case 0656).

\begin{table}[t]
\centering
\small
\begin{tabular}{lrr}
\toprule
Configuration & CSA (\%) & $\Delta$ (pp) \\
\midrule
First-round image only & 53.4 & -- \\
Best-of-2 resampling & 55.0 & +1.6 \\
Generic-feedback repair & 56.6 & +3.2 \\
Localized repair w/o locking & 57.8 & +4.4 \\
\textbf{Localized repair (ImageRSI)} & \textbf{65.7} & \textbf{+12.3} \\
\bottomrule
\end{tabular}
\caption{Second-round strategies on the test set (431 requests). All strategies share the first-round image and use at most two images per request; $\Delta$ is the gain over the first-round image only.}
\label{tab:repair_ablation}
\end{table}

\subsection{Experience Library Size}
\label{app:experience_size}

Table~\ref{tab:experience_size} lists the values plotted in Figure~\ref{fig:ablation}b. The four libraries are nested prefixes of the same fixed random order over the experience set, and all other settings follow localized repair.

\begin{table}[t]
\centering
\small
\begin{tabular}{lrrr}
\toprule
Library & Entries & CSA (\%) & $\Delta$ (pp) \\
\midrule
0\% (no experience) & 0 & 55.7 & -- \\
25\% & 131 & 56.8 & +1.2 \\
50\% & 266 & 61.3 & +5.6 \\
\textbf{100\%} & \textbf{527} & \textbf{65.7} & \textbf{+10.0} \\
\bottomrule
\end{tabular}
\caption{CSA on the test set with nested experience libraries; $\Delta$ is the gain over using no experience.}
\label{tab:experience_size}
\end{table}

\section{Supplementary Figures}
\label{app:supp_figures}

\begin{figure}[t]
\centering
\includegraphics[width=\columnwidth]{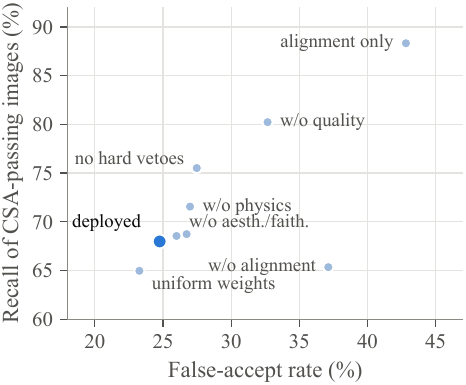}
\caption{Offline ablation of the five-dimensional judge on delivered ImageReq-Bench images. Each point re-aggregates the recorded dimension scores under one configuration and compares the resulting decision with the external CSA label. Removing the alignment dimension raises the false-accept rate from 24.8\% to 37.1\%; removing the quality dimension or the hard vetoes increases recall but also the false-accept rate; removing the low-confidence discount changes no decision and thus coincides with the deployed configuration. ``w/o aesth./faith.'' marks the two nearby points that remove the aesthetics and faithfulness dimensions, respectively.}
\label{fig:judge_ablation}
\end{figure}

Figure~\ref{fig:judge_ablation} reports an offline ablation of the judge's score aggregation.

\end{document}